\documentclass[10pt,twocolumn,letterpaper]{article}

\usepackage{cvpr}              
\usepackage{xcolor}
\usepackage{stfloats}   

\usepackage{multirow}
\usepackage[table]{xcolor} 
\usepackage{graphicx}
\usepackage{tabularx}
\usepackage{array}
\usepackage{adjustbox}
\usepackage{tabularx}
\usepackage{makecell}
\usepackage[normalem]{ulem}
\usepackage{colortbl}

\definecolor{cvprblue}{rgb}{0.21,0.49,0.74}
\usepackage[pagebackref,breaklinks,colorlinks,allcolors=cvprblue]{hyperref}

\def\confName{CVPR}
\def\confYear{2026}
\def\name{VidPrism} 
\title{\textit{\name}: Heterogeneous Mixture of Experts for Image-to-Video Transfer}

\author{Rui Lin \quad Chuanming Wang \quad Huadong Ma\textsuperscript{\thanks{Corresponding author: Huadong Ma}}\\
    State Key Laboratory of Networking and Switching Technology,\\
    Beijing University of Posts and Telecommunications, China\\
    \textit{\{lr\_507, wcm, mhd\}@bupt.edu.cn}
}

\begin{document}
\maketitle
\begin{abstract}
With the rapid development of pre-training technologies, adapting large-scale Vision-Language Models (VLMs) for video understanding \emph{\ie} image-to-video transfer learning has become a dominant paradigm. 
To achieve superior performance, it raises as an effective strategy among recent advances to employ Mixture-of-Experts (MoE) to enhance VLMs' temporal modeling capabilities. However, conventional MoE designs suffer from expert homogenization, where all experts act as identical generalists, inefficiently learning spatio-temporal features from undifferentiated video streams.
To overcome this problem, we propose \name, a novel heterogeneous temporal Mixture-of-Experts framework.
{\name} pioneers a division of labor by deploying functionally specialized experts, each assuming a role ranging from spatial understanding to temporal modeling.
To feed these specialists appropriately, we introduce a content-aware, multi-rate sampling module that dynamically generates streams ranging from semantically rich to motion-focused representations, providing specialized inputs for experts.
Furthermore, a dynamic, bidirectional fusion mechanism enables synergistic information exchange between these pathways, leading to a comprehensive video representation. 
Extensive experiments on various video recognition benchmarks demonstrate that {\name} achieves state-of-the-art performance and effectively fosters expert specialization.
Our source code is available at \href{https://github.com/Lrrrr549/VidPrism.git}{https://github.com/Lrrrr549/VidPrism.git}. 
\end{abstract}    
\section{Introduction}
\label{sec:intro}

The emergence of large-scale fundamental models has driven a paradigm shift in the field of artificial intelligence, where Visual Language Models (VLMs)~\cite{clip,videoclip,viclip} are at the forefront of this revolution. These models are pre-trained on massive image-text pair datasets and demonstrate superior few-shot learning capabilities in a variety of downstream vision tasks. Therefore, significant performance improvements have been achieved in few-shot image analysis~\cite{Bar_2024_CVPR,Guo_2025_CVPR}. However, compared to images, video data exhibits stronger continuity and temporal relevance, and the widely used image pre-training process results in video language models lacking the ability to handle temporal relevance. This weakens their performance in video understanding, which has been more widely used in real-world applications.

\begin{figure} 
  \centering
  \includegraphics[width=1\linewidth]{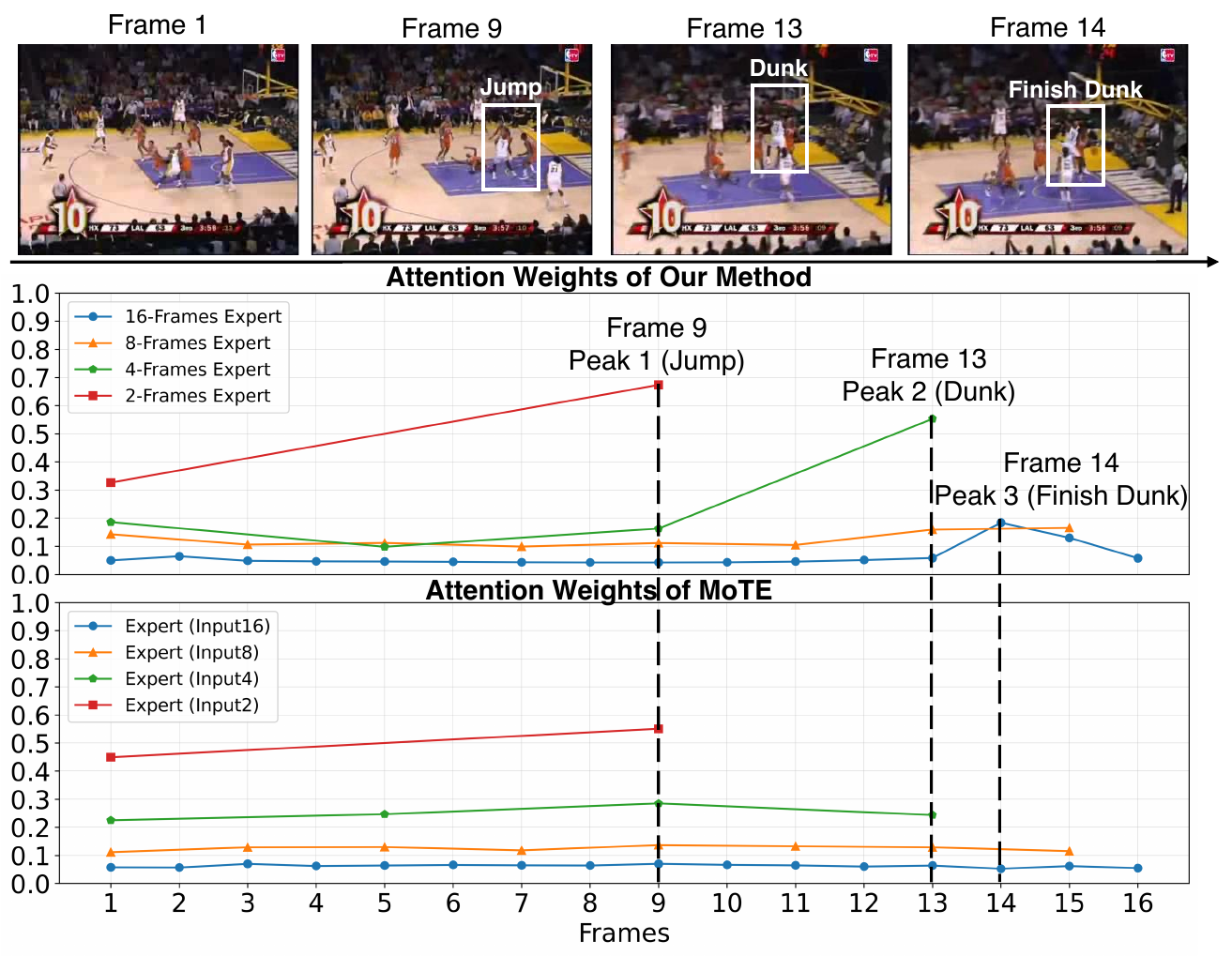} 
  \caption{\textbf{Visualization of inter-frame attention distribution.} We select a video from the Dunk category in UCF‑101~\cite{ucf101} to compare inter‑frame attention between MoTE and our method. Using the same four-frame configurations for fairness, our model exhibits clear attention peaks at key moments,
   while the homogeneous MoTE baseline shows flat, indistinct distributions. 
   }
  \label{fig:fig1}
    \vspace{-2em}
\end{figure}
Consequently, how to adapt these VLMs to the domain of video understanding (\emph{i.e.} Image-to-Video (I2V) transfer learning) has become one of the core issues in current research. Among recent advance works, adding additional temporal modules~\cite{xclip,text4vis,efficient} to explicitly model of temporal dynamics has attracted increasing attention due to its flexibility and strong performance, and some works have begun to explore the introduction of Mixture-of-Experts (MoE) architecture as such temporal modules~\cite{mote}.

Nevertheless, traditional MoE designs face an inherent limitation when applied to complex and multifaceted data such as video: expert homogenization. In standard MoE, all experts are trained using the same undifferentiated input stream. This forces each expert to become a generalist, redundantly learning overlapping features. This design is inefficient for video because it fails to create specialized computational paths to model inherently different types of information, such as the static content of a scene and its temporal evolution. 
The limitation is verified in Figure~\ref{fig:fig1}. The baseline MoTE~\cite{mote} method exhibits flat and diffused attention, failing to identify critical moments in the action.

This problem raises us a question intuitively, can we process the video frames by a group of heterogeneous experts, each operating at different perspectives. This insight also conforms to the core concept in \emph{two-stream hypothesis} of neuroscience~\cite{goodale1992separate}, which posits parallel ‘what’(spatial) and ‘how’(temporal) pathways in the brain to understand complex video contents.
However, constructing such a heterogeneous multipath architecture still faces two key challenges: 1. How to provide the most relevant input to each expert? 
2. How can these experts achieve effective collaboration and information sharing?

To address these challenges, we first extend this concept of the two-stream hypothesis from a fixed dual-pathway architecture to a more flexible multipathway framework, and develop a novel heterogeneous temporal MoE framework called {\name}. {\name} enables specialized processing of information at different temporal scales by dynamically generating specialized multi-rate input streams and binding them to heterogeneous expert networks. It incorporates a group of spatial-temporal experts focusing on different spatial-temporal resolutions, thereby achieving greater knowledge capacity and finer specialization within each functional domain.

To provide specialized input for experts (\underline{challenge 1}), {\name} introduces a content-aware mechanism that splits the input video into multiple distinct multi-rate streams. The low-rate stream captures keyframes with rich semantic context and routes them to the expert group responsible for spatial reasoning. Simultaneously, the high-rate stream focuses on high-frequency temporal variations and directs them to the expert group specifically responsible for motion modeling. To facilitate collaboration and information exchange among experts (\underline{challenge 2}), we design a bidirectional fusion module, which helps the model to build a holistic understanding of the video by integrating global scene context and complex temporal dynamics. Finally, the outputs of all experts are aggregated to form a unified video-level representation for classification. 
{\name} allows for a dynamic division of labor among functionally heterogeneous experts, enabling the model to focus on critical action moments, as validated in Figure~\ref{fig:fig1}.

Our main contributions can be summarized as follows:
\begin{itemize}
\item We are the first to propose a heterogeneous Mixture-of-Experts architecture for image-to-video transfer, where expert networks are explicitly specialized and bound to different temporal scales. 
\item We design a content-aware multi-rate input generation module that dynamically generates semantically rich slow-rate streams and motion-dense fast-rate streams, providing specialized input for different experts.
\item We introduce a dynamic, bidirectional fusion mechanism that enables synergistic information exchange between the spatial and temporal expert pathways. 
\end{itemize}
\section{Related Work}
\label{sec:relatedwork}

\noindent
\textbf{Video Recognition.} Early video recognition relied on handcrafted features~\cite{iDT}, combined appearance and motion via optical flow. Two-Stream Networks~\cite{2stream,conv2stream} processed RGB frames and optical flow separately to model motion explicitly. 3D CNNs~\cite{k400,tube,tsm,R3D,temporal} learned spatio-temporal representations directly from raw pixels through spatial-temporal convolutions, reducing dependence on pre-computed motion. Transformer-based models~\cite{timesformer,vivit,multiscale,evl,vswint,xclip,aim} treated videos as sequences of spatio-temporal patches, using self-attention to capture long-range dependencies beyond CNNs’ local receptive fields. Recently, multimodal large models~\cite{step3,glm,internvideo2,qwen3} integrate visual and textual modalities into unified representations, advancing video understanding abilities. 

\noindent
\textbf{I2V Transfer Learning.} Adapting pre-trained image foundation models, particularly VLMs~\cite{clip, viclip, videoclip}, for video recognition has become a prevalent and parameter-efficient paradigm. To bridge the domain gap from static images to dynamic videos, early work augmented frozen image encoders with simple temporal pooling. More sophisticated approaches followed, introducing lightweight, trainable modules like temporal adapters~\cite{stadapter, evl} and embeddings~\cite{actionclip, xclip} to explicitly model temporal dependencies. This trend has evolved towards more intricate structural modifications, including learnable queries~\cite{focusvideo} and advanced spatio-temporal adapters~\cite{efficient}. 

\noindent
\textbf{Multi‑Pathway Temporal Modeling.} 
{Multi‑pathway architectures~\cite{slowfast} employ dual‑rate pathways to capture slow spatial semantics and fast motion, inspiring a range of spatial‑temporal decomposition strategies~\cite{k400,R3D,tsm} and feature‑level hierarchies~\cite{TPN,CFAD,TPP}. Extensions to multimodal large language models~\cite{sfmllm,maaz2024videogpt+,xu2024sfllava,xu2025sfllava1.5,wang2024grounded} have further shown the promise of multi‑path designs for long‑range and fine‑grained reasoning. Furthermore, approaches employing Mixture‑of‑Experts (MoE) frameworks for temporal modeling~\cite{mote} introduce conditional expert routing to enhance temporal specialization. 
Yet, most existing multi‑path methods operate at fixed, preset frame rates. This static design limits adaptability. 
In contrast, our heterogeneous MoE dynamically associates specialized experts with content‑adaptive, multi‑rate inputs, enabling fine‑grained collaboration across temporal scales. 
}


\begin{figure*}
    \centering
    \includegraphics[width=0.9\textwidth]{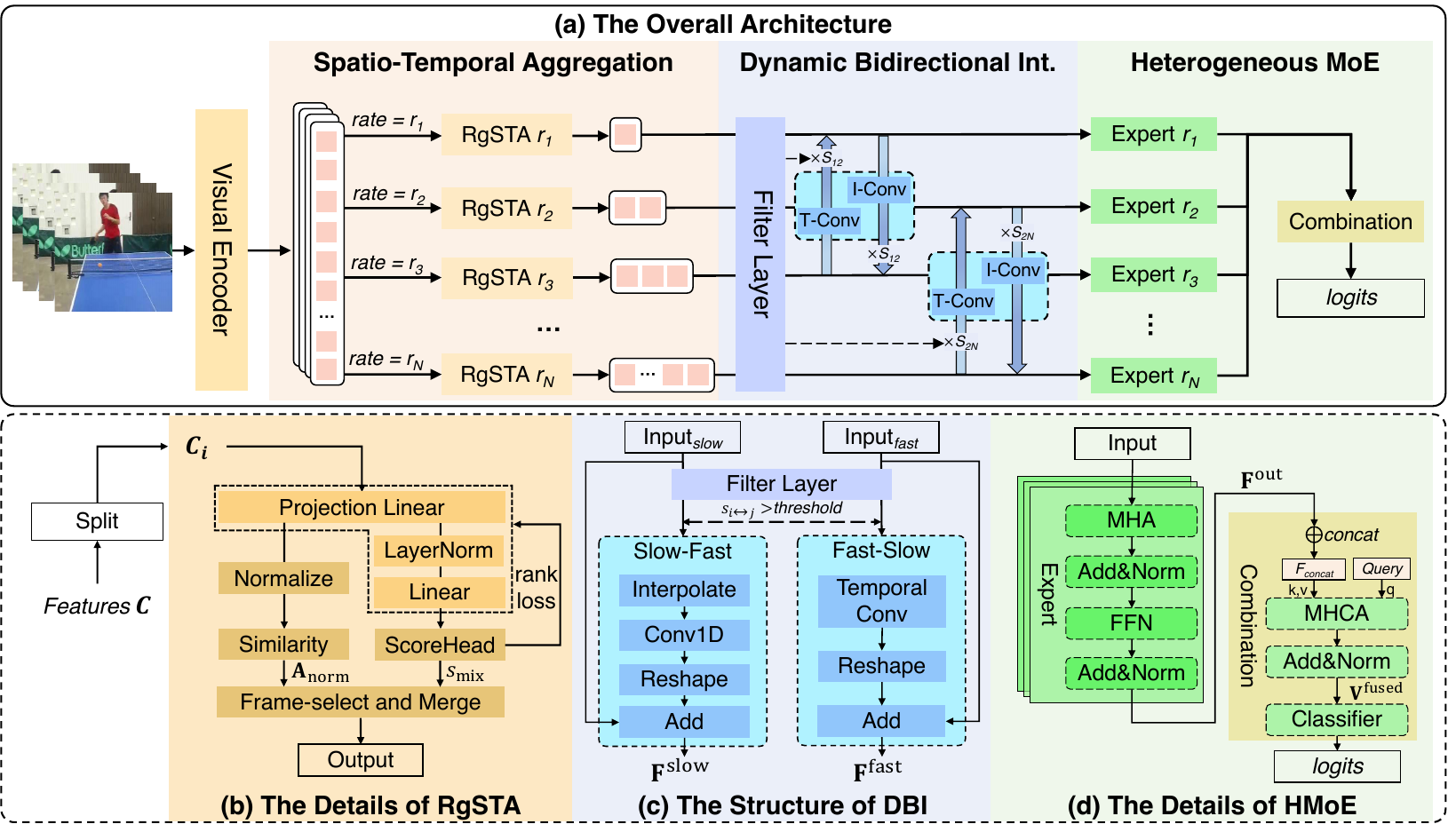}\vspace{-0.8em}
    \caption{
        An overview of {\name}:
        (1) A visual encoder processes the input video to extract a sequence of frame-level features. 
        (2) {Multiple RgSTA modules} then tokenizes these features into $N$ parallel streams, each with a distinct spatio-temporal resolution. 
        (3) {The DBI module} facilitates selective information fusion across the multi-rate streams. 
        (4) HMoE models each enriched stream and integrates outputs into a comprehensive global representation for final classification.
        }\vspace{-1.5em}
    \label{fig:overview}
\end{figure*}

\section{Method}

The overview of our method is shown in Figure~\ref{fig:overview}. Next, we will introduce the specific design of {\name} in detail.



\subsection{{Diverse }Spatio-Temporal Aggregation}  \label{merge}

{Given an input video, we first send it to a visual encoder to extract frame-level features $\mathbf{C} \in \mathbb{R}^{T \times B \times D}$}, where$T$, $B$, and$D$ denote the number of frames, batch size, and feature dimension, respectively. Since input video usually contains redundant information, spatial-temporal aggregation should be performed to produce compact and robust representations. Traditional strategy samples frame features at a fixed rate, which may cause the overlook of important temporal information. Thus, we propose a diverse spatio-temporal aggregation method, in which frame features are sampled at different rates and processed by specialized pathways.

The core designed module in the aggragation step is Rate-guided Spatio-Temporal Aggregation Module (RgSTA), where rate stands for different spatio-temporal resolutions, aiming at efficiently fusing temporal features while maximising the retention of critical dynamic spatio-temporal information by merging rather than discarding information. The details of STA is shown in the Figure~\ref{fig:overview}(b). After spliting the {frame features $\mathbf{C} $ according to the rate $r_i$ into $\mathbf{c}_i \in \mathbb{R}^{T_i \times B \times D}$, where $T_i = T / r_i$ denotes the temporal length of the $i$‑th pathway}, RgSTA first assesses the importance of each frame feature, and a hybrid scoring mechanism is introduced, which will combine learnable importance scores $s_{\text{pred}}$ and the intrinsic properties $s_{\text{norm}}$ of the features themselves, {aiming to achieve a more comprehensive frame importance evaluation.}

For the scores $s_{\text{pred}}$, we introduce a score head to predict the importance of each feature. Specifically, the input feature $\mathbf{c}_i $ is first projected into a metric space, and then a scalar score is obtained through the scoring header. {The process can be formulated as:
\vspace{-0.5ex}
\begin{align}
    s_{\text{pred},i} = \text{ScoreHead}(\text{LN}(\text{MetricProj}(\mathbf{c}_i)),
\end{align}
where $\operatorname{LN}$ means LayerNorm\cite{ba2016layer}.} Both MetricProj and ScoreHead are implemented by a linear layer.
This design allows the model to autonomously learn to determine which features are critical based on task goals.

For the intrinsic {properties} $s_{\text{norm}}$, we are of the view that the L2 norm of a feature can reflect its intrinsic importance to some extent, i.e., features with larger paradigms usually contain richer {signals~\cite{larger,ye2018rethinking}}. Therefore, we compute the L2 norm of each feature as its intrinsic score $s_{\text{norm},i} = ||\mathbf{c}_i||_2$.
The final importance score $s_{\text{mix}}$ is a weighted fusion of the above two outputs:
\begin{align}
    s_{\text{mix},i} = \alpha \cdot s_{\text{pred},i} + (1 - \alpha) \cdot s_{\text{norm},i}
\end{align}
where $\alpha \in [0, 1]$ is a hyperparameter used to balance the importance between $s_{\text{pred}}$ and $s_{\text{norm}}$.

Within each group, we select a feature with the highest score $s_{\text{mix}}$ to form a kept set, while the remaining features constitute a rest set. To avoid discarding the information of the rest set, we design a fusion mechanism. {We compute the cosine similarity between features in the rest set $\mathbf{C}_{\text{rest}} \in \mathbb{R}^{1\times B\times D}$ and the kept set $\mathbf{C}_{\text{kept}} \in \mathbb{R}^{(T_i-1)\times B\times D}$ to obtain a similarity matrix $\mathbf{S}_{\text{rest}\to\text{kept}} \in \mathbb{R}^{(T_i-1)\times 1\times B}$.}

This matrix is then normalized by a Softmax function to yield the weight matrix $\mathbf{A}_{\text{norm}}$:
\begin{align}
    \mathbf{Z}_{\text{kept}} &= \text{Norm}(\text{MetricProj}(\mathbf{C}_{\text{kept}})), \\
    \mathbf{Z}_{\text{rest}} &= \text{Norm}(\text{MetricProj}(\mathbf{C}_{\text{rest}})), \\
    \mathbf{S}_{\text{rest}\to\text{kept}} &= \mathbf{Z}_{\text{rest}} \mathbf{Z}_{\text{kept}}^T, \\
    \mathbf{A}_{\text{norm}} &= \text{Softmax}\!\left(\frac{\mathbf{S}_{\text{rest}\to\text{kept}}}{\tau}\right),
\end{align}
where $\tau$ is the temperature coefficient. For each feature in the rest set, its information is distributed to the kept set based on these attention weights. Specifically, the information from the rest set is accumulated into the kept set by a weighted summation:
\begin{align}
    \mathbf{C}'_{\text{kept}} = \mathbf{C}_{\text{kept}} + \delta (\mathbf{A}_{\text{norm}})^T \mathbf{C}_{\text{rest}},
\end{align}
where $\delta$ is a scaling factor to moderate the magnitude of the update. The feature $\mathbf{F} \in  \mathbf{C}'_{\text{kept}}$ generated by each RgSTA subsequently enter parallel processing pathways. 

This approach yields two benefits. On the one hand, it makes the sequence more compact by selecting only the most important frames. On the other hand, instead of discarding information from the dropped frames, it integrates features into the selected keyframes. This process enriches the keyframes, ensuring that no valuable information is lost while making the representation more efficient.

\begin{figure}
    \centering
    \includegraphics[width=\columnwidth]{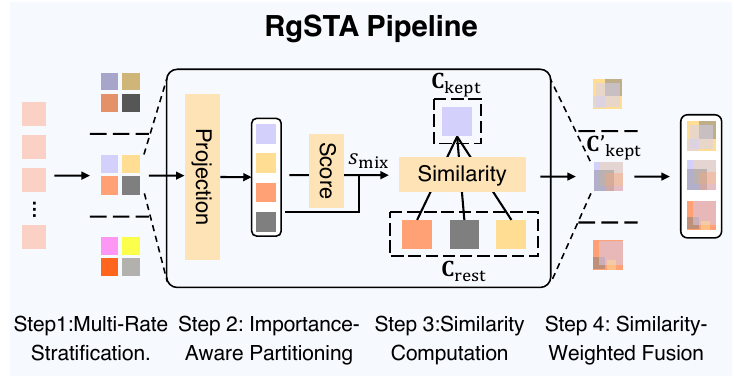}
    \caption{
    A pipeline of STAM: (1) Input features are stratified into groups. (2) Within each group, an importance score divides the features into kept and rest sets. (3) Pairwise similarity between the two sets is computed to generate attention weights. (4) Ultimately, the features in the rest set are aggregated back to the kept set by attentional weighted summation.
        }\vspace{-1em}
    \label{fig:pipeline}
\end{figure}

\subsection{Dynamic Bidirectional Interaction} \label{interaction}

In \name, different pathways capture the dynamic information of video at different temporal resolutions. To enable these specialised pathways to collaborate rather than operate in isolation, we propose to adopt a dynamic bidirectional interaction scheme to exchange information among these pathways. According to Figure~\ref{fig:overview}(c), by employing a filter mechanism, selective bidirectional information is exchanged by the designed Dynamic Bidirectional Interaction (DBI) module.

To determine the necessity and strength of information exchange between any two pathways $i$ and $j$, we design a lightweight filter network $\mathcal{G}_{i \leftrightarrow j}$ for each pair of pathways $(i, j)$. This network dynamically generates an interaction score $s_{i \leftrightarrow j} \in [0, 1]$ based on the global context information of the two pathways.

Specifically, given $\mathbf{F}_i \in \mathbb{R}^{T_i \times D}$ for the $i$-th pathway and $\mathbf{F}_j \in \mathbb{R}^{T_j \times D}$ for the $j$-th pathway, we first extract the respective global summary vectors $\mathbf{g}_i$ and $\mathbf{g}_j$ by time dimensional global average pooling. These two vectors are then spliced and fed into a proprietary MLP, which is activated by a Sigmoid function to obtain the interaction score:
\vspace{-1ex}
\begin{align}
    \mathbf{g}_k &= \frac{1}{T_k} \sum_{t=1}^{T_k} \mathbf{F}_k, \quad k \in \{i, j\} \\
    s_{i \leftrightarrow j} &= \sigma\left(\text{MLP}_{i \leftrightarrow j}([\mathbf{g}_i; \mathbf{g}_j])\right),
\end{align}
where $[\cdot; \cdot]$ denotes a vector splice and $\sigma$ is a Sigmoid function. $s_{i \leftrightarrow j}$ acts as a gate score that dynamically regulates the amount of information flowing between pathways $i$ and $j$. {To improve computational efficiency and promote sparse connectivity, we set a threshold. 
When $s_{i \leftrightarrow j}$ meets the threshold, the bidirectional interaction is conducted by DBI modules.} DBI is performed on two different-rate pathways, so it supports two complementary streams: a Slow-to-Fast information flow providing high-level context, and a Fast-to-Slow information flow delivering rich temporal details.

\textbf{Slow-to-Fast Path.} When information needs to be passed from the slow pathway to the fast pathway, the alignment of the temporal and channel dimensions must be addressed. We first use linear interpolation $\mathbf{I}$ to upsample the temporal dimension $T_i$ of the slow pathway feature $\mathbf{F}_i$ to align with the temporal dimension $T_j$ of the fast pathway feature $\mathbf{F}_j$ to obtain $\mathbf{F}'_i$. The $\mathbf{F}'_i$ is then passed through a $1 \times 1$ convolutional layer for channel adaptation and feature transformation. The transformed features are then multiplied by a dynamically gated fraction $s_{i \leftrightarrow j}$ and added to the original features $\mathbf{F}_j$ of the fast pathway in a residual-join fashion. This can be formulated as:
\begin{align}
    \mathbf{F}_j^{\text{fast}} = \mathbf{F}_j + s_{i \leftrightarrow j} \cdot \text{Conv}_{1 \times 1}(\mathbf{I}(\mathbf{F}_i)).
\end{align}

\textbf{Fast-to-Slow Path.} When information flows from the fast to the slow pathway, then temporal downsampling is required. We apply a temporal convolution with a step size of $S = R_j / R_i$ to the fast pathway feature $\mathbf{F}_i$, where $R_i$ and $R_j$ are the temporal sampling rates of pathways $i$ and $j$, respectively. This operation reduces the temporal resolution while effectively aggregating dynamic information within a local time window. The convolved features are similarly multiplied by the corresponding gating scores $s_{i \leftrightarrow j}$, added to the slow pathway features $\mathbf{F}_j$:
\begin{align}
    \mathbf{F}_j^{\text{slow}} = \mathbf{F}_j + s_{i \leftrightarrow j} \cdot \text{T-Conv}(\mathbf{F}_i).
\end{align}
In this way, information about movement changes captured by the fast pathway can be refined and injected into the slow pathway, thus enriching its understanding of the overall structure of the movement. When one-to-multi information exchange occurs, we process raw features from all paths in parallel, weight them by their gating scores, and accumulate the aggregated information into the target path.


\subsection{Heterogeneous Mixture-of-Experts}

The spatio-temporal features processed by~\ref{merge} and~\ref{interaction} possess differing lengths and semantic focus. To better capture the information within them and integrate them into a globally unified video representation, as shown in Figure~\ref{fig:overview}(d), we design a Heterogeneous Mixture-of-Experts (HMoE) module comprising a set of heterogeneous experts and a combination block.

\textbf{Heterogeneous Experts.} Each input feature corresponds to a separate expert module. These expert modules are designed to learn and model the internal temporal dependencies at the corresponding time scales, thereby achieving functional heterogeneity. Each expert is designed as a standard Transformer layer and is exclusively trained on a single, specific temporal rate. Given an input sequence $\mathbf{F}_i \in \mathbb{R}^{T_i \times D}$, the processing flow is as follows:
\begin{align}
    \mathbf{F}'_i &= \text{LN}(\mathbf{F}_i + \text{MHSA}(\mathbf{F}_i)), \\
    \mathbf{F}^{\text{out}}_i &= \text{LN}(\mathbf{F}'_i + \text{FFN}(\mathbf{F}'_i)).
\end{align}
Among them, MHSA refers to Multi-Head Self-Attention, FFN is a feed-forward network containing two linear layers and a GELU activation function.

\textbf{Combination Mechanism.} After experts process their respective sequences independently, a combination module is adopted to aggregate this diverse information. The module introduces a learnable global query vector that makes it traverses the outputs of all experts and extracts the comprehensive information that is most important for the task.

We splice the output sequences $\mathbf{F}^{\text{out}}_1, \dots, \mathbf{F}^{\text{out}}_N$ of all the $N$ experts in the time dimension to obtain $\mathbf{F}_{\text{concat}} \in  \mathbb{R}^{(\sum T_i) \times D}$. Also, we introduce a learnable global query parameter $\mathbf{q}_{\text{global}} \in \mathbb{R}^{1 \times D}$, whose task is to learn how to efficiently summarise information from $\mathbf{F}_{\text{concat}}$ during training.
We perform a multi-headed cross-attention operation with $\mathbf{q}_{\text{global}}$ as the Query and $\mathbf{F}_{\text{concat}}$ as the Key and Value:
\vspace{-1ex}
\begin{align}
\mathbf{V}_{\text{fused}} &= \text{MHA}(\mathbf{q}_{\text{global}}, \mathbf{F}_{\text{concat}}), \\
{logits} &= \text{Classifier}(\mathbf{V}_{\text{fused}}),
\end{align}
where MHA is Multi-Head Attention. The output of this operation $\mathbf{V}_{\text{fused}} \in \mathbb{R}^{1 \times D}$ is a single vector representing a weighted aggregation of all experts' features, with the weights determined dynamically by the similarity of the query vector to the individual features. Finally, $\mathbf{V}_{\text{fused}}$ passes through a classifier to produce classification logits.

\subsection{Training Objectives}

To monitor the learning process of the {\name}, we employ four loss functions.

\textbf{Classification Loss.} We employ the standard classification cross-entropy loss $\mathcal{L}_\text{cls}$, applied to the final fused feature vector $\mathbf{v}_{\text{fused}}$. This loss serves as the primary guidance for the network's overall learning.
\begin{equation}
 \mathcal{L}_{\text{cls}} 
 = - \frac{1}{N} \sum_{i=1}^{N} \sum_{c=1}^{C} 
 y_{i,c}\, \log\!\left( \text{softmax}(\text{logits}_{i,c}) \right),
\end{equation}
where $N$ is the batch size, $C$ is the number of classes, $y_{i,c}$ is the one-hot ground-truth label, and $\text{logits}_{i,c}$ is the predicted score for class $c$ of the $i$-th sample.

\textbf{Ranking Loss.} To guide the ScoreHead module (in \ref{merge}) towards learning more reliable importance rankings, we introduce an auxiliary ranking loss. This loss aims to make the distribution of predicted scores $s_{\text{pred}}$ approximate the distribution of target scores $s_{\text{tgt}}$. $s_{\text{tgt}}$ is computed by combining feature norm and average similarity within its window:
\vspace{-1.5ex}
\begin{align}
s_{\text{tgt},i} = ||\mathbf{c}_i||_2 + \frac{1}{N} \sum_{j=1}^{N} \mathbf{S}_{ij}.
\end{align}

\vspace{-1.5ex}
We employ the Kullback-Leibler divergence to measure the difference between these two score distributions, defining it as the ranking loss $\mathcal{L}_{\text{rank}}$:
\begin{align}
\mathcal{L}_{\text{rank}} = \text{KL}(\text{Softmax}(s_{\text{tgt}} / T_s) \ || \text{Softmax}(s_{\text{pred}} / T_s)),
\end{align}
where $T_s$ is a temperature hyperparameter. 

\textbf{Diversity Loss.} Concurrently, to encourage experts to learn complementary features, we introduce the diversity loss $\mathcal{L}_{\text{div}}$. This loss aims to maximise the distance between feature outputs from different experts. Specifically, it computes the cosine similarity between output features $\bar{\mathbf{F}}^{\text{out}}_i$ and $\bar{\mathbf{F}}^{\text{out}}_j$ for each expert pair $(i, j)$:
\begin{align}
    \mathcal{L}_{\text{div}} = \frac{1}{\binom{N}{2}} \sum_{i<j} \text{D}(\bar{\mathbf{F}}^{\text{out}}_i, \bar{\mathbf{F}}^{\text{out}}_j).
\end{align}

\vspace{-1ex}
By minimising this loss , we incentivise each expert to focus on capturing dynamic information at its unique temporal scale, thereby enhancing the representational capability of the entire hybrid expert system.

\textbf{Gate Balancing Loss.} To prevent the Readout module from over-relying on a minority of experts during fusion while neglecting others' contributions, we introduce the gating balancing loss $\mathcal{L}_{\text{gate}}$. 
Given the gated weight matrix $\mathbf{W} \in \mathbb{R}^{B \times N}$ produced by the Readout module's attention, the balancing loss is defined as:
\vspace{-1ex}
\begin{align}
    C_i &= \frac{1}{B} \sum _{b=1}^{B} W_{b,i}, \\
    \mathcal{L}_{\text{gate}} &= N \cdot \sum_{i=1}^{N} C_i^2,
\end{align}
where $W_{b,i}$ is expert $i$'s contribution to sample $b$, and $C_i$ is each expert's average contribution within the batch. This loss takes effect by penalising imbalanced contributions.

The model's overall loss function is a weighted sum of the aforementioned losses, defined as follows:
\begin{align}
    \mathcal{L}_{\text{total}} = \mathcal{L}_{\text{cls}} + \lambda_{\text{rank}}\mathcal{L}_{\text{rank}} + \lambda_{\text{div}}\mathcal{L}_{\text{div}} + \lambda_{\text{gate}}\mathcal{L}_{\text{gate}}
\end{align}
where $\lambda_{\text{rank}}$, $\lambda_{\text{div}}$, and $\lambda_{\text{gate}}$ are hyperparameters used to balance the contributions of each loss component.
\begin{table}[!t]
    \centering
    \caption{Performance comparison on K400. Per-view GFLOPs are reported. Inputs mean frames$\times$crops$\times$clips.}
    \label{tab:k400}
    \renewcommand{\arraystretch}{1.0}
    \small
    \begin{tabularx}{\linewidth}{lcXXc}
        \toprule
        \textbf{Method} & \textbf{Inputs} & \textbf{Top-1(\%)} & \textbf{Top-5(\%)} & \textbf{GFLOPs} \\
        \hline
        \rowcolor{blue!10} \multicolumn{5}{c}{Adapting} \\
        ST-Adapter-B/16~\cite{stadapter}               & 8$\times$1$\times$3 & 82.0 & 95.7 & 148 \\
        ST-Adapter-B/16~\cite{stadapter}               & 32$\times$1$\times$3 & 82.7 & 96.2 & 607 \\
        AIM-B/16~\cite{aim}                            & 8$\times$1$\times$3 & 83.9 & 96.3 & 202 \\
        ActionCLIP-B/16~\cite{actionclip}              & 32$\times$10$\times$3 & 83.8 & 96.2 & 563 \\
        X-CLIP-B/16~\cite{xclip}                       & 8$\times$4$\times$3 & 83.8 & 96.7 & 145 \\
        Vita-CLIP-B/16~\cite{vitaclip}                 & 16$\times$4$\times$3 & 82.9 & 96.3 & 190 \\
        STAN-conv-B/16~\cite{stan}                     & 8$\times$1$\times$3 & 83.1 & 96.0 & 238 \\
        M$^{2}$-CLIP-B/16~\cite{wang2024m2clip}        & 8$\times$4$\times$3 & 83.4 & 96.3 & 214 \\
        M$^{2}$-CLIP-B/16~\cite{wang2024m2clip}        & 32$\times$4$\times$3 & 84.1 & 96.8 & 842 \\
        ETL-ViCLIP-B/16~\cite{efficient}               & 8$\times$4$\times$3 & 82.2 & 96.2  & --  \\
        FocusVideo-B/16~\cite{focusvideo}              & 8$\times$4$\times$3 & 84.1 & 96.5 & 204 \\
        FocusVideo-B/16~\cite{focusvideo}              & 32$\times$4$\times$3 & 84.7 & 96.8 & 816 \\
        \rowcolor{yellow!10} \multicolumn{5}{c}{Tuning} \\
        MoTE-B/16~\cite{mote}                          & 8$\times$4$\times$3 & 83.0 & 96.3 & 141 \\
        OST-B/16~\cite{ost}                            & 16$\times$1$\times$1 & 83.2 & --   & --  \\
        \hline
        \name-B/16                                    & 8$\times$4$\times$3 & 84.0 & 96.8 & 162\\
        \name-B/16                                    & 32$\times$4$\times$3 & \textbf{85.1} & \textbf{97.0} & 721 \\
        \hline
        \rowcolor{blue!10} \multicolumn{5}{c}{Adapting} \\
        ST-Adapter-L/14~\cite{stadapter}               & 8$\times$1$\times$3 & 86.7 & 97.5 & 687 \\
        ST-Adapter-L/14~\cite{stadapter}               & 32$\times$1$\times$3 & 87.2 & 97.6 & 2749 \\
        AIM-L/14~\cite{aim}                            & 8$\times$1$\times$3 & 86.8 & 97.2 & 934 \\
        DUALPATH-L/14~\cite{dual_trans}                & 32$\times$1$\times$3 & 87.7 & 97.8 & --   \\
        M$^{2}$-CLIP-L/14~\cite{wang2024m2clip}        & 32$\times$4$\times$3 & 87.0 & 97.6 & --   \\
        FocusVideo-L/14~\cite{focusvideo}              & 8$\times$4$\times$3 & 87.2 & 97.7 & 914 \\
        FocusVideo-L/14~\cite{focusvideo}              & 32$\times$4$\times$3 & 88.0 & \textbf{97.9} & 3656 \\
        \rowcolor{yellow!10} \multicolumn{5}{c}{Tuning} \\
        Text4Vis-L/14~\cite{text4vis}                  & 32$\times$4$\times$3 & 87.1 & 97.4 & 1662 \\
        MoTE-L/14~\cite{mote}                          & 8$\times$4$\times$3 & 86.8 & 97.5 & 649 \\
        MoTE-L/14~\cite{mote}                          & 16$\times$4$\times$3 & 87.2 & 97.7 & 1299 \\
        \hline
        \name-L/14                                    & 8$\times$4$\times$3 & 87.4 & 97.8 & 632 \\ 
        \name-L/14                                    & 32$\times$4$\times$3 & \textbf{88.6} & \textbf{97.9} &  2594 \\ 
        \bottomrule
    \end{tabularx}
    \vspace{-1em}
\end{table}

\section{Experiments}

\subsection{Experimental Setup}
\noindent
We conduct experiments on four video benchmarks: Kinetics-400~\cite{k400}, UCF-101~\cite{ucf101}, HMDB-51~\cite{hmdb} and SomethingSomething V2 (SSv2)~\cite{ssv2}. Our research involves few-shot and fully-supervised video recognition. We use CLIP~\cite{clip} with ViT-B/16 and ViT-L/14 as the visual backbone. The sparse frame sampling strategy for the video uses 8 and 32 frames during training and inference. For 8-frame sampling, we choose a triple expert with rates of 2,4,8; for 32-frame sampling, we use a quadruple expert architecture with rates of 2,4,8,16. All experiments are conducted on two NVIDIA A100 GPUs. 

\begin{table*}[!t]
\centering
\caption{Performance comparison with state-of-the-art methods on HMDB51, UCF101 and SSv2 under various $K$-shot settings. In particular, the visual backbone of \name-C is CLIP, while \name-M uses VideoMAEv2. }
\label{tab:fewshot}
\renewcommand{\arraystretch}{1}
{
\begin{tabular}{lcccccccccccc}
\toprule
\multirow{2}{*}{Method}
& \multicolumn{4}{c}{\textbf{HMDB51}}
& \multicolumn{4}{c}{\textbf{UCF101}}
& \multicolumn{4}{c}{\textbf{SSv2}} \\
\cline{2-5}
\cline{6-9}
\cline{10-13}
& $K$=2 & $K$=4 & $K$=8 & $K$=16
& $K$=2 & $K$=4 & $K$=8 & $K$=16
& $K$=2 & $K$=4 & $K$=8 & $K$=16 \\
\hline
\rowcolor{blue!10} \multicolumn{13}{c}{Zero-shot} \\
CLIP~\cite{clip}                & 37.2 & 37.2 & 37.2 & 37.2 & 62.8 & 62.8 & 62.8 & 62.8 & 2.8 & 2.8 & 2.8 & 2.8 \\
ViCLIP~\cite{viclip}            & 47.8 & 47.8 & 47.8 & 47.8 & 71.0 & 71.0 & 71.0 & 71.0 & 5.1 & 5.1 & 5.1 & 5.1 \\
\rowcolor{yellow!10} \multicolumn{13}{c}{Adapting} \\
A5~\cite{A5}                    & 39.7 & 50.7 & 56.0 & 62.4 & 71.4 & 79.9 & 85.7 & 89.9 & 4.4 & 5.1 & 6.1 & 9.7 \\
X-CLIP~\cite{xclip}             & 53.0 & 57.3 & 62.8 & 64.0 & 76.4 & 83.4 & 88.3 & 91.4 & 3.9 & 4.5 & 6.8 & 10.6 \\
ActionCLIP~\cite{actionclip}    & 47.5 & 57.9 & 57.7 & 63.2 & 70.0 & 71.5 & 73.0 & 91.4 & 4.4 & 5.3 & 8.4 & 11.1 \\
ViFi-CLIP~\cite{vificlip}       & 57.2 & 62.7 & 64.5 & 66.8 & 80.7 & 85.1 & 90.0 & 92.7 & 6.2 & 7.4 & 8.5 & 12.4 \\
ETL~\cite{efficient}            & 61.2 & 62.3 & 67.1 & 70.4 & 86.1 & 90.2 & 92.7 & 94.8 & \textbf{9.1} & \textbf{10.4} & 13.4 & 17.5 \\
\rowcolor{green!10} \multicolumn{13}{c}{Tuning} \\
MAXI~\cite{vificlip}            & 58.0 & 60.1 & 65.0 & 66.5 & 86.8 & 89.3 & 92.4 & 93.5 & 7.3 & 7.4 & 8.4 & 12.4 \\
VideoMAEv2~\cite{videomae}      & 38.6 & 52.3 & 63.2 & 72.2 & 73.4 & 86.8 & 92.1 & 94.7 & 5.0 & 8.6 & 12.2 & 17.1 \\
OST~\cite{ost}                  & 60.0 & 62.5 & 65.6 & 67.3 & 83.0 & 88.4 & 91.3 & 93.9 & 7.3 & 8.4 & 8.5 & 11.5 \\
MoTE~\cite{mote}                & \textbf{61.3} & 63.9 & 67.2 & 68.2 & 88.8 & 91.0 & 92.3 & 93.6 & 7.3 & 8.5 & 9.5 & 12.2 \\
ViCLIP~\cite{viclip}            & 53.7 & 60.4 & 64.5 & 70.3 & 83.0 & 88.0 & 92.1 & 93.2 & 8.7 & 9.7 & 11.6 & 15.4 \\
\hline
\name-C                         & 55.0 & \textbf{64.1} & \textbf{67.6} & \textbf{74.1} & 88.1 & \textbf{91.6} & \textbf{93.8} & \textbf{96.0} & 4.9 & 7.3 & 9.3 & 14.1 \\
\name-M                         & 53.4 & 63.3 & \textbf{68.3} & \textbf{73.2} & \textbf{94.3} & \textbf{95.6} & \textbf{95.9} & \textbf{96.6} & 6.4 & 9.5 & \textbf{13.6} & \textbf{18.3} \\

\bottomrule
\end{tabular}
}\vspace{-1em}
\end{table*}

\subsection{Main Results}

\noindent
\textbf{Close-set Results.} Table \ref{tab:k400} shows the results of the comparison between {\name} and the current video model of SOTA on the K400 dataset. We can first observe that our proposed {\name} framework demonstrates significant performance benefits over the MoTE baseline, achieving a 1.0\% accuracy improvement with a ViT-B/16 backbone and 8-frame input. This improvement persists with the ViT-L/14 backbone, maintaining a 0.6\% advantage while requiring nearly identical computational resources. These results provide strong evidence that setting up dedicated timing experts is a more effective and efficient strategy for video representation learning. Second, using an 8-frame input with a ViT-B/16 backbone, {\name} delivers 84.0\% accuracy, nearly identical to the 84.1\% from FocusVideo. Notably, our method operates at a much lower computational cost, consuming just 162 GFLOPs while FocusVideo demands 204 GFLOPs. Furthermore, when the input is extended to 32 frames, {\name} achieves a Top-1 accuracy of 85.1\%, outperforming state-of-the-art methods such as MoTE and FocusVideo. This demonstrates that {\name} achieves an excellent performance-efficiency balance, proving the effectiveness of our heterogeneous MoE architecture. When using the larger ViT-L/14 architecture, {\name} achieves further performance gains, proving its scalability and viability in larger architectures. The above results demonstrate {\name}'s efficiency and robustness in K400 video tasks.

\noindent
\textbf{Few-shot video recognition.} Table \ref{tab:fewshot} provides the evaluation of our method on the few-shot video recognition task. In few-shot experiments, in addition to the original CLIP, we additionally used VideoMAEv2 as the visual backbone to test the scalability of our architecture across different visual backbones. The model based on CLIP is called \name-C, while the model based on VideoMAE is called \name-M. On HMDB-51, our \name-C model establishes new state-of-the-art results, achieving accuracies of 64.1\%, 67.6\%, and 74.1\% for $K$=4, 8, and 16, respectively. 
On UCF-101, our \name-M variant achieves a clean sweep, setting new state-of-the-art results in all few-shot scenarios and peaking at 96.6\% for $K$=16. The model further pushes the performance boundary on the SSv2 dataset, reaching 13.6\% for $K$=8 and a new record of 18.3\% for $K$=16.
These results span multiple datasets and $K$-shot settings, demonstrating the adaptability of our framework.

\begin{table}[t]
	\centering
    \renewcommand{\arraystretch}{1}
	\caption{Ablation study on Number of Experts.}
	\label{tab:experts}
		\begin{tabular}{cccc}
			\toprule
			\makecell{Experts\\Numbers} & \makecell{Experts\\Rates} & UCF-101 & HMDB-51 \\
			\midrule
			\multirow{4}{*}{1} 
			& 2   & 94.8 & 74.1 \\
			& 4   & 94.8 & 74.3 \\
			& 8   & 94.5 & 74.2 \\
			& 16  & 94.6 & 74.2 \\
			\hline
			\multirow{6}{*}{2} 
			& 2,4  & 95.1 & 74.9 \\
			& 4,8  & 95.2 & 73.7 \\
			& 8,16 & 95.2 & 74.8 \\
			& 2,8  & 95.3 & 73.9 \\
			& 4,16 & 95.2 & 74.5 \\
			& 2,16 & 95.1 & 73.9 \\
			\hline
			\multirow{3}{*}{3} 
			& 2,4,8  & 94.6 & 74.2 \\
			& 4,8,16 & 94.3 & 74.1 \\
			& 2, 4, 16 & 94.1 & 74.2 \\
			\hline
			\rowcolor{gray!10} \textbf{4} & \textbf{2, 4, 8, 16} & \textbf{95.9} & \textbf{76.3} \\
			\bottomrule
		\end{tabular}
        \vspace{-1.5em}
\end{table}

\begin{table}
    \centering
    \renewcommand{\arraystretch}{0.9}
    \caption{Effect of Different Supervision.}
    \label{tab:loss}
    \begin{tabular}{ccc}
        \toprule
        Type & UCF-101 & HMDB-51 \\
        \midrule
        $\mathcal{L}_\mathrm{cls}$ & 94.5 &  75.1 \\
        $\mathcal{L}_\mathrm{cls} + \mathcal{L}_\mathrm{rank}$ & 95.2 & 75.6 \\
        $\mathcal{L}_\mathrm{cls} + \mathcal{L}_\mathrm{rank} +\mathcal{L}_\mathrm{div}$ & 95.4 & 75.5 \\
        \rowcolor{gray!10} \textbf{$\mathcal{L}_\mathrm{cls} + \mathcal{L}_\mathrm{rank} +\mathcal{L}_\mathrm{div} +\mathcal{L}_\mathrm{gate}$} & \textbf{95.9} & \textbf{76.3} \\
        \bottomrule
    \end{tabular}
    \vspace{-1em}
\end{table}

\subsection{Ablation Studies}

To evaluate the effectiveness of our proposed individual modules, we execute detailed ablation experiments on UCF101 and HMDB51. Unless otherwise stated, all experiments in this section use CLIP ViT-B/16 as the backbone network with 32 input frames. In tables, default settings are in \colorbox{gray!10}{gray}, best results are in \textbf{bold}.

\textbf{Varying Number of Experts.} We investigate the impact of varying the number of experts and their corresponding time-rate configurations, as shown in Table~\ref{tab:experts}. Starting with a single expert as the baseline, akin to a single-path model, we achieve 94.8\% accuracy on UCF-101 and 74.3\% on HMDB-51, highlighting the effectiveness yet limitations of a single timescale. Increasing the number of experts to two allowed modeling of two time scales, improving performance; for instance, the rate combination {2, 8} reached 95.3\% on UCF-101. However, using three experts cause a slight performance drop, suggesting that more experts alone can introduce redundancy or optimization challenges. The best performance, 95.9\% on UCF-101 and 76.3\% on HMDB-51, is achieved with four experts covering rates {2, 4, 8, 16}, demonstrating the value of diverse temporal resolutions. Based on these results, we select this four-expert configuration for our final model.

\noindent
\textbf{Different Aggregation Strategies.} We briefly try four different methods for feature aggregation at different time resolutions, as shown in Table~\ref{tab:aggr}. In addition to the original STAM, we have also selected direct sampling, average pooling, and max pooling. STAM perform best among several methods, demonstrating that the module can better extract information from the input video stream.

\noindent
\textbf{Direction of Interaction.} In addition to the bidirectional interaction (Double), we test three additional functional interaction modules. As shown in Table~\ref{tab:comb}, ‘None’ indicates that this module is not used, whereas ‘Slow2Fast’ and ‘Fast2Slow’ specify a unidirectional fusion from the slow to the fast pathway and from the fast to the slow pathway, respectively. The results demonstrate that the bidirectional interaction mechanism achieves the best performance among the evaluated methods. In contrast, unidirectional interaction limits the comprehensiveness of the features and can be detrimental to the final representation.

\noindent
\textbf{Combination Strategies.} In Table~\ref{tab:comb}, we compare five combination methods, ranging from simple functions (Avg Pooling, Linear, MLP) to attention mechanisms. While Local Attention processes each expert's output in isolation, Global Attention captures dependencies across all experts. The results confirm that GlobalAttn is the superior approach, achieving state-of-the-art accuracies of 95.9\% on UCF101 and 76.3\% on HMDB51. This highlights the critical need to model global, long-range relationships for effective feature fusion, a capability that localized or context-agnostic methods lack.

\begin{table*}[htbp]
	\centering
	\caption{Ablation studies on the three modules of {\name}.}
	\footnotesize
    \renewcommand{\arraystretch}{0.9}
	\begin{subtable}{0.32\textwidth}
		\centering
		\begin{tabular}{ccc}
			\toprule
			Type & UCF-101 & HMDB-51 \\
			\midrule
			Hard Sampling & 94.7 & 75.0 \\
			Avg Pooling & 95.1 & 75.2 \\
			Max Pooling & 95.2 & 75.5 \\
            \rowcolor{gray!10} \textbf{RgSTA} & \textbf{95.9} & \textbf{76.3} \\
        \bottomrule
		\end{tabular}
		\caption{Aggration}
        \label{tab:aggr}
	\end{subtable}
	\hfill
	\begin{subtable}{0.32\textwidth}
		\centering
		\begin{tabular}{ccc}
			\toprule
			Type & UCF-101 & HMDB-51 \\
			\midrule
			None & 95.3 & 74.2\\
			Slow2Fast & 95.2 & 73.6 \\
			Fast2Slow & 95.0 & 74.9 \\
			\rowcolor{gray!10} \textbf{DBI} & \textbf{95.9} & \textbf{76.3} \\
			\bottomrule
		\end{tabular}
		\caption{Interaction}
        \label{tab:inter}
	\end{subtable}
	\hfill
	\begin{subtable}{0.32\textwidth}
		\centering
		\begin{tabular}{ccc}
			\toprule
			Type & UCF-101 & HMDB-51 \\
			\midrule
			Avg Pooling & 94.8 & 75.6 \\
			Linear & 94.5 & 73.9 \\
			MLP & 94.7 & 74.1\\
			LocalAttn & 95.1 & 74.6 \\
			\rowcolor{gray!10} \textbf{GlobalAttn} & \textbf{95.9} & \textbf{76.3} \\
			\bottomrule
		\end{tabular}
		\caption{Combination}
        \label{tab:comb}
	\end{subtable}
    \vspace{-2.5em}
\end{table*}

\noindent
\textbf{Effect of Different Supervision.} We conduct ablation experiments in Table~\ref{tab:loss} to evaluate the contributions of each component in the supervision strategy. Starting from the baseline using only cross-entropy (CE) loss, adding ranking loss result in improvements of 0.7\% and 0.5\% on UCF-101 and HMDB-51, respectively. Further addition of diversity loss continuously improve model performance by promoting functional differentiation among experts. Finally, the introduction of gating loss yield the most significant gain, enabling the complete model to achieve 95.9\% and 76.3\% accuracy on UCF-101 and HMDB-51, respectively. These results demonstrate the necessity of each supervision module and showcase their synergistic effect.

\subsection{Visualization}
\begin{figure}[t]
    \centering
    \includegraphics[width=\columnwidth]{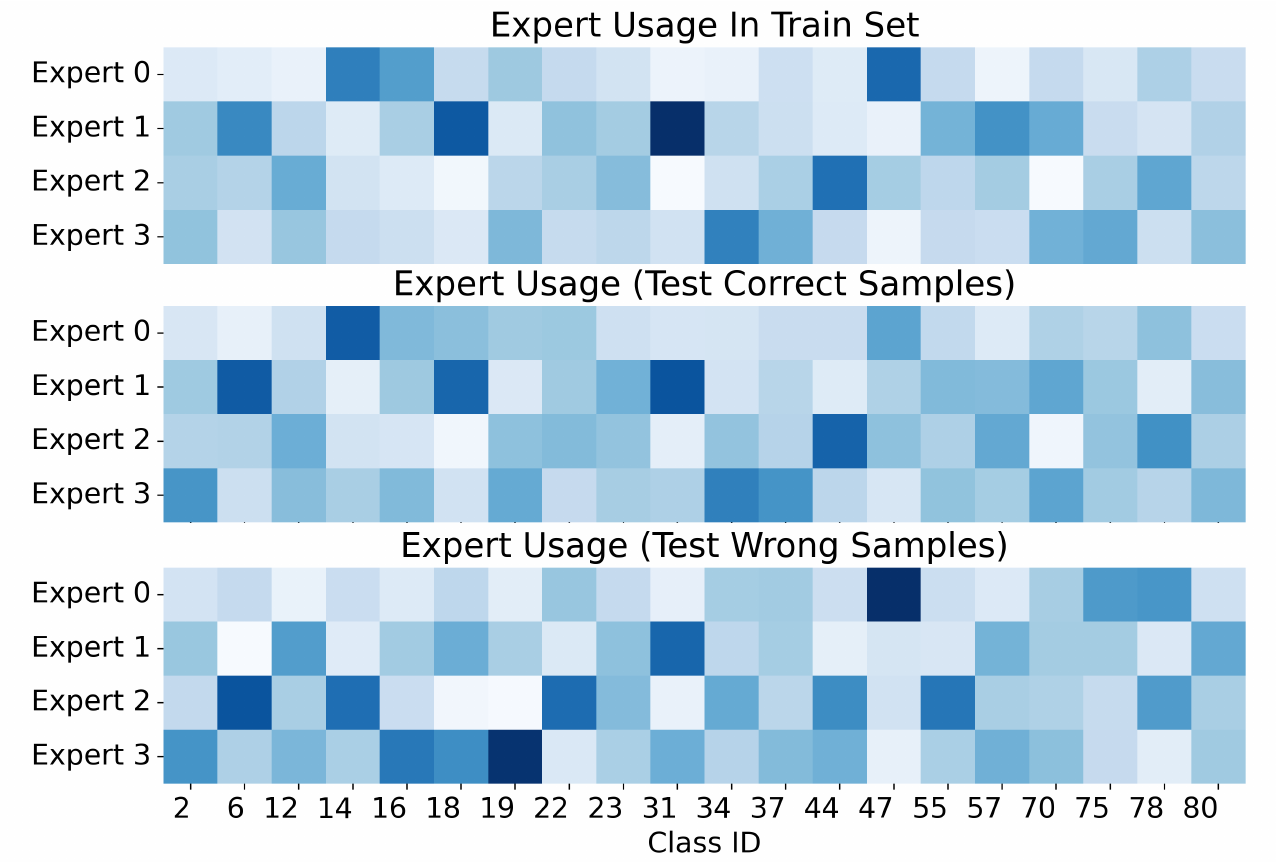}
    \caption{Visualization of Expert Usage on Training and Test Sets.}
    \label{fig:expert_usage}
    \vspace{-1.5em}
\end{figure}

We visualise expert usage on the UCF-101 dataset in Figure~\ref{fig:expert_usage}, separately aggregating training samples, correctly classified test samples, and misclassified samples.
The activation patterns of the training set and correctly classified samples are highly consistent, indicating that the model learns a stable and generalisable expert assignment strategy; moreover, different action categories exhibit distinct expert usage, evidencing a clear functional division of labour among heterogeneous experts rather than redundant, interchangeable units.
In contrast, the heatmaps of misclassified samples deviate markedly from these patterns, for example, errors in category 16 are associated with over-activation of expert 3, suggesting that incorrect predictions often arise when model fails to choose the most appropriate expert.

\vspace{0em}
\section{Conclusion}

In this paper, we introduce {\name}, a novel heterogeneous temporal Mixture-of-Experts framework designed to advance image-to-video transfer learning. We address the limitation of expert homogenization in conventional MoE models by pioneering a functional division of labor. Through a content-aware, multi-rate sampling module, we provide specialized data streams to experts dedicated to distinct functions. A dynamic, bidirectional fusion mechanism enriches these specialized representations. Extensive experiments demonstrate that {\name} not only achieves state-of-the-art performance on major video recognition benchmarks but also effectively fosters  expert specialization. 

\section{Acknowledgment}
This work was supported in part by the National Natural Science Foundation of China under Grant U24B20176 and 62406038.

{
    \small
    \bibliographystyle{ieeenat_fullname}
    \bibliography{main}
}


\end{document}